\documentclass{article}

\usepackage{microtype}
\usepackage{graphicx}
\usepackage{booktabs} %
\usepackage[table]{xcolor}

\usepackage{hyperref}
\usepackage{subcaption}

\usepackage{icml2025}

\usepackage{amsmath}
\usepackage{amssymb}
\usepackage{mathtools}
\usepackage{amsthm}

\usepackage[capitalize,noabbrev]{cleveref}

\theoremstyle{plain}
\newtheorem{theorem}{Theorem}[section]

\theoremstyle{definition}

\theoremstyle{remark}
\newtheorem{remark}[theorem]{Remark}

\usepackage[textsize=tiny]{todonotes}

\colorlet{lgreen}{green!10}
\colorlet{lblue}{blue!10}
\colorlet{lred}{red!10}

\usepackage{tikz}
\NewDocumentCommand{\progressbar}{m O{2} O{blue!70}}{%
  \begin{tikzpicture}
    \fill[gray!30] (0,0) rectangle (#2,0.3); %
    \fill[#3] (0,0) rectangle ({#1/100*#2},0.3); %
  \end{tikzpicture}%
}

\usepackage{listings}

\definecolor{Numbers}{RGB}{0,137,84}      %
\definecolor{Definitions}{RGB}{0,0,255}   %
\definecolor{Functions}{RGB}{126,93,23}   %
\definecolor{Variables}{RGB}{0,17,134}    %
\definecolor{Comments}{RGB}{0,100,0}      %
\definecolor{Strings}{RGB}{178,0,2}       %
\definecolor{Operators}{RGB}{0,0,0}       %
\definecolor{BaseFunctions}{RGB}{192,0,227} %
\definecolor{Bakgroung}{RGB}{255,255,255} %

\lstdefinelanguage{PythonVSCodeLight}{
    language=Python,
    morekeywords={
        torch, nn, autograd, optim, utils, cuda, device, Tensor,
        tensor, from_numpy, zeros, ones, randn, manual_seed, no_grad, save, load,
        Linear, Conv2d, ReLU, Sigmoid, Tanh, Dropout, Sequential, Module,
        SGD, Adam, lr_scheduler,
        CrossEntropyLoss, MSELoss, BCELoss, NLLLoss,
    },
    keywords=[2]{abs, norm, kthvalue, where},
    keywordstyle=[2]\color{Functions}, %
    sensitive=true
}

\icmltitlerunning{$\mu$-MoE: Test-Time Pruning as Micro-Grained Mixture-of-Experts}

\begin{document}

\twocolumn[
\icmltitle{$\boldsymbol\mu$-MoE: Test-Time Pruning as Micro-Grained Mixture-of-Experts}

\icmlsetsymbol{equal}{*}

\begin{icmlauthorlist}
\icmlauthor{Toshiaki Koike-Akino}{merl}
\icmlauthor{Jing Liu}{merl}
\icmlauthor{Ye Wang}{merl}
\end{icmlauthorlist}

\icmlaffiliation{merl}{Mitsubishi Electric Research Laboratories (MERL), 201 Broadway, Cambridge, MA 02139, USA.}

\icmlcorrespondingauthor{Toshiaki Koike-Akino}{koike@merl.com}

\icmlkeywords{Machine Learning, Mixture of Experts, Dynamic Network, Activation-Aware Pruning, LLM}

\vskip 0.3in
]

\printAffiliationsAndNotice{}  %

\begin{abstract}
To tackle the huge computational demand of large foundation models, activation-aware compression techniques without retraining have been introduced. 
However, since these rely on calibration data, domain shift may arise for unknown downstream tasks.
With a computationally efficient calibration, activation-aware pruning can be executed for every prompt adaptively, yet achieving reduced complexity at inference. 
We formulate it as a mixture of micro-experts, called $\mu$-MoE. 
Several experiments demonstrate that $\mu$-MoE can dynamically adapt to task/prompt-dependent structured sparsity on the fly.
\end{abstract}

\section{Introduction}

Large foundation models~\cite{touvron2023llama, achiam2023gpt,liu2023LLaVA} have shown excellent performance across a variety
of general tasks~\cite{wei2022emergent,katz2024gpt, bubeck2023sparks}. 
Nonetheless, these models, with billions of parameters, demand significant computational resources~\cite{schwartz2020green}. 
Towards increasing the accessibility of large language models (LLMs), a number of compression methods~\cite{xu2023survey, zhu2024survey, bai2024beyond} have been introduced: e.g., partial activation~\cite{jiang2024mixtral, lin2024moe}, pruning~\cite{frantar2023sparsegpt, sun2023simple, bai2024sparsellm, hassibi1993optimal},
quantization~\cite{frantar2022gptq, lin2024awq, wang2024q}, knowledge distillation~\cite{hsieh2023distilling, deepseekai2025deepseekr1incentivizingreasoningcapability, hwang2024pc}, and rank reduction~\cite{yuan2023asvd, liu2024deepseek, hwang2024pc, saxena2024eigen}.

Test-time scaling~\cite{chen2024expanding, muennighoff2025s1} is a paradigm to improve LLM performance by increasing inference computation. 
We instead consider extra test-time computing to reduce the total cost of inference on the fly. 
Specifically, we use a pruning operation that dynamically selects important weights depending on each prompt.
We view it as a mixture of micro-experts, namely $\mu$-MoE, where, instead of a few massive experts, we may have a massive number of single-parameter weight multiplier experts.
We show that activation-aware pruning makes this concept feasible.
The contributions of this paper are summarized below:
\begin{itemize}
  \setlength{\itemsep}{2pt}
  \setlength{\parskip}{0pt}
  \setlength{\parsep}{0pt}
    \item We propose a mixture of micro-experts $\mu$-MoE concept to realize the finest-grained adaptation.
    \item We adopt low-complexity activation-aware pruning to realize test-time LLM compression as $\mu$-MoE.
    \item We tackle the domain shift issue caused by offline calibration required for baseline static pruning.
    \item We demonstrate the benefit of $\mu$-MoE over state-of-the-art methods for several LLM benchmarks.
\end{itemize}

\begin{figure*}[t]
    \centering
    \includegraphics[width=0.7\linewidth]{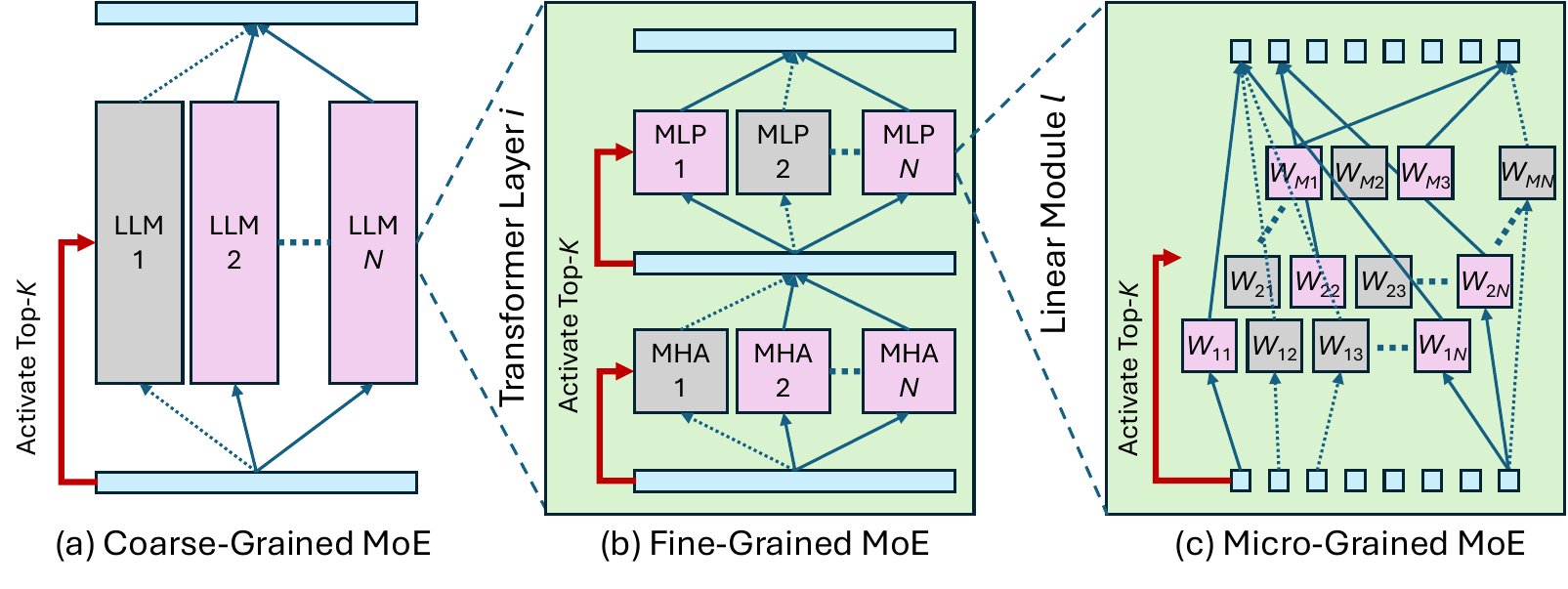}
    \vskip-2ex
    \caption{Coarse to micro-grained MoE.}
    \label{fig:moe}
\end{figure*}

\begin{figure}[t]
    \centering
    \includegraphics[width=\linewidth]{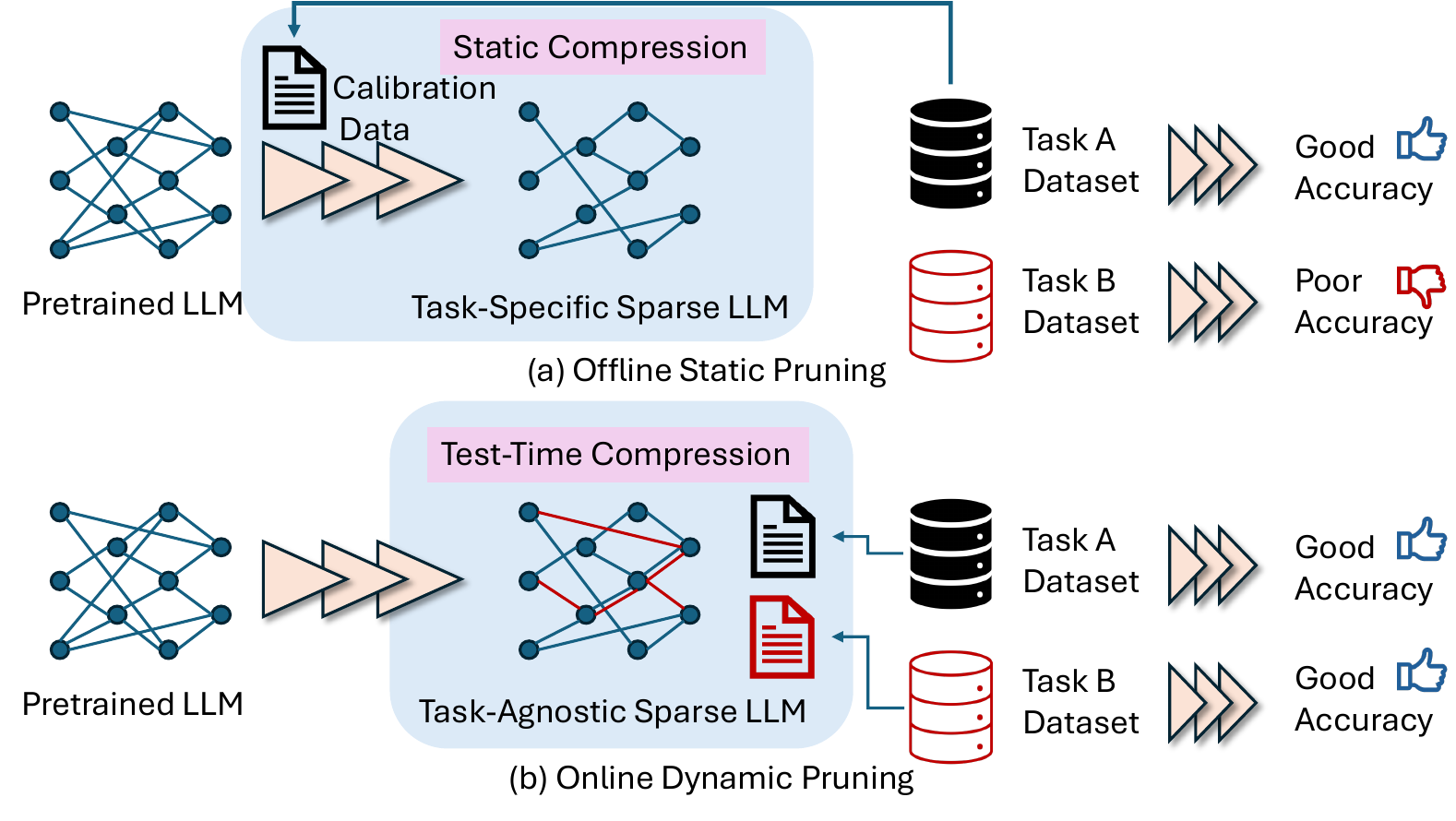}
    \vskip-2ex
    \caption{Offline vs online pruning: dynamic pruning finds prompt-dependent sparse structure at test time, preventing domain shift.}
    \label{fig:prune}
\end{figure}

\section{Micro-Grained Mixture of Experts (MoE)}

\paragraph{Coarse to Micro-Grained MoE}

\cref{fig:moe} illustrates the MoE framework from coarse-grained to micro-grained scales.
A coarse-grained MoE may involve multiple LLM modules that activate depending on provided prompts.
Most MoEs use mid- to fine-grained architectures.
For example, Mixtral-8x7B~\cite{jiang2024mixtral} has 8 multi-layer perceptrons (MLPs) per layer, but select only 2 of them, realizing a six-fold speedup at inference.
Finest-grained experts would be a single-parameter weight multiplier within the linear modules of LLMs.
We consider such a mixture of micro-experts, referred to as $\mu$-MoE.

$\mu$-MoE employs test-time adaptation to reduce the total test-time compute, i.e., online dynamic pruning to reduce the number of active weights for inference computation.
Besides the computational efficiency, the online dynamic pruning may potentially solve the domain shift issue caused by mismatched calibration data used for activation-aware offline static pruning as illustrated in \cref{fig:prune}.

\paragraph{Activation-Aware Pruning}

As an alternative to magnitude-based pruning, activation-aware pruning~\cite{williams2023impact} leverages the statistics of the activation features.
Let $X\in\mathbb{R}^{d\times T}$ be input activation of embedding dimension $d$ for token length $T$.
The aim is to minimize the approximation loss:
\begin{align}
    \mathcal{L} &=
    \mathbb{E}_{X}
    \big[
    \big\| (W - \hat{W}) X \big\|^2
    \big],
    \label{eq:loss}
\end{align}
where $W\in\mathbb{R}^{d'\times d}$ is a weight matrix and $\hat{W}$ is its pruned version such that only a fraction $\rho$ of the weights are active: $\|\hat{W}\|_0 = \rho\cdot dd'$.
SparseGPT~\cite{frantar2023sparsegpt} is such an activation-aware LLM pruning inspired by optimal brain surgeon~\cite{hassibi1993optimal}.
It uses a scoring metric for pruning:
\begin{align}
S_{i,j} &=
|W_{i,j}|^2 / \big[ \mathsf{Chol}[ (\bar{X} \bar{X}^\top + \lambda I)^{-1} ] \big]_{j,j}^2,
\end{align}
where $\mathsf{Chol}[\cdot]$ denotes the Cholesky factorization, $\bar{X}\in\mathbb{R}^{d\times T_\mathrm{c}}$ are $T_\mathrm{c}$ tokens of input activation features sampled from calibration data.
Here $\lambda$ is a small damping factor.
Although SparseGPT achieves good pruning performance, the computational cost to obtain the sparse matrix is at least of cubic order due to the inverse Hessian calculation: $\mathcal{O}[d^3 + d d' T_\mathrm{c}]$.
In addition, it needs extra computations to update the non-zero weights with the Gaussian elimination.

Wanda~\cite{sun2023simple} simplifies the metric by approximating with a diagonal correlation: $\bar{X} \bar{X}^\top +\lambda I \simeq \mathsf{diag}[\bar{X} \bar{X}^\top]$ .
The modified score is written as:
\begin{align}
    S'_{i,j} &=
    |W_{i,j}| \cdot \| \bar{X}_{j,:} \|_2.
\end{align}
It only activates weights within the top-$\rho$ fraction in this metric.
The pseudo code in PyTorch is given below:
\begin{lstlisting}
# W:(d',d), X:(d,Tc), kc=int((1-rho) * d)
S = W.abs() * X.norm(p=2, dim=-1)  
val, _ = torch.kthvalue(S, dim=-1, k=kc)
W = torch.where(S > val[:,None], W, 0)
\end{lstlisting}
Wanda requires only quadratic complexity $\mathcal{O}[3dd' + dT_\mathrm{c}]$ including norm calculation, metric product, top-$k$ search, and comparators, yet achieves performance competitive with SparseGPT.
It yields semi-structured sparsity with a constant number of active weights per row.

\begin{remark}
    While the original Wanda uses {torch.sort}, this sorting complexity of $\mathcal{O}[d'd\log(d)]$ can be reduced by torch.topk or torch.kthvalue. We note that torch.kthvalue has a linear theoretical complexity. 
    See \cref{app:wanda}.
\end{remark}

\paragraph{Instant Wanda Pruning as $\mu$-MoE}

To realize $\mu$-MoE, we use test-time tokens $X$ as an online calibration to prune weights, rather than offline calibration tokens $\bar{X}$.
When the number of active weights is reduced, the inference complexity of linear modules will be reduced from $\mathcal{O}[dd' T]$ to $\mathcal{O}[\rho dd' T]$. 
However, it is meaningless if the cost to find the top-$\rho$ weights is higher than the reduced complexity.
Hence, SparseGPT is not suitable due to its cubic complexity, while Wanda is a viable candidate.
The total complexity with online Wanda pruning is $\mathcal{O}[3dd'+dT+\rho dd' T]$.
The complexity ratio (compared to the original $\mathcal{O}[dd'T]$) is on the order of:
\begin{align}
\frac{3dd'+dT+\rho dd' T}{dd' T} &=
\rho + \frac{3}{T} + \frac{1}{d'} \simeq \rho, \quad (T, d'\gg 1).
\notag
\end{align}
This suggests that instant Wanda pruning for every prompt has almost no additional complexity compared to the original full-weight operations for a large enough token length ($T\gg 1$) and embedding dimension ($d'\gg 1$).
Moreover, Wanda is known to be robust even with a single calibration sample~\cite{williams2023impact}, motivating us to use test-time Wanda pruning for our $\mu$-MoE.

\begin{table*}[h]
\centering
\caption{Perplexity ($\downarrow$) of OPT models with different pruning methods at $60$--$40\%$ active weights. 
Red-highlighted cells indicate that Wanda uses a matched calibration-test dataset.
Bold-face letters indicate the best cases.
}
\label{tab:perp_opt}
\small
\setlength{\tabcolsep}{4pt}
\newcolumntype{g}{>{\columncolor{lblue}}r}
\begin{tabular}{l rrrg rrrg rrrg}
  \toprule
  Active Weights & \multicolumn{4}{c}{60\%} & \multicolumn{4}{c}{50\%} & \multicolumn{4}{c}{40\%} 
  \\
  \cmidrule(lr){2-5}
  \cmidrule(lr){6-9}
  \cmidrule(lr){10-13}  
  Test Dataset & WT2 & PTB & C4 & Avg & WT2 & PTB & C4 & Avg & WT2 & PTB & C4 & Avg \\
  \midrule
  \multicolumn{13}{c}{OPT-125M (WT2: 27.7, PTB: 39.0, C4: 26.6, Avg: 31.1)}
  \\
  \midrule
  Magnitude Prune & 
  43.9 & 71.9 & 39.8 &  51.9 & %
  90.9 & 168.6 & 71.9 & 110.4 &  %
  533.2 & 906.3 & 349.9 & 596.5 %
  \\
  Wanda (WT2 Calib)  & 
  \cellcolor{lred} 30.8 & 44.2 & 30.3 & 35.1 & 
  \cellcolor{lred} 37.1 & 56.3 & 37.5 & 43.6 & 
  \cellcolor{lred} 65.4 & 106.5 & 37.5 & 81.6  
  \\
  Wanda (PTB Calib)  & 
  32.4 & \cellcolor{lred}43.8 & 31.7 & 36.0 & 
  44.0 & \cellcolor{lred}52.8 & 42.1 & 46.3 & 
  89.7 & \cellcolor{lred}\bf{86.5} & 87.6 & 87.9 
  \\
  Wanda (C4 Calib) &
  30.7 & 44.4 & \cellcolor{lred}29.3 & 34.8 & 
  39.1 & 57.1 & \cellcolor{lred}34.8 & 43.7 & 
  75.1 & 104.3 & \cellcolor{lred}60.4 & 80.0  
  \\
  \rowcolor{lgreen}
  \bf{$\boldsymbol\mu$-MoE} &
  \bf{30.3} & \bf{43.3} &  \bf{28.6} & \bf{34.1} & 
  \bf{35.8} & \bf{51.8} &  \bf{32.5} & \bf{40.1} & 
  \bf{61.0} & {87.5} &  \bf{52.3} & \bf{66.9}  
  \\
  \toprule
  \multicolumn{13}{c}{OPT-1.3B (WT2: 14.6, PTB: 20.3, C4: 16.1, Avg: 17.0)}
  \\
  \midrule
  Magnitude Prune & 
  150.0 & 306.1 & 103.2 &  186.4 & %
  799.5 & 1438.1 & 298.5 & 845.4 &  %
  6218.7 & 7303.5 & 2385.8 & 5302.7 %
  \\
  Wanda (WT2 Calib)  & 
  \cellcolor{lred} 16.6 & 23.6 & 18.8 & 19.6 & 
  \cellcolor{lred} 18.9 & 28.3 & 22.4 & 23.2 & 
  \cellcolor{lred} 25.6 & 43.2 & 34.1 & 34.3  
  \\
  Wanda (PTB Calib)  & 
  16.4 & \cellcolor{lred}22.5 & 19.6 & 19.5 & 
  20.4 & \cellcolor{lred}25.3 & 25.9 & 23.9 & 
  34.6 & \cellcolor{lred}34.0 & 48.6 & 39.1 
  \\
  Wanda (C4 Calib) &
  \bf{16.0} & 23.4 &\cellcolor{lred} 18.0 & 19.1 & 
  18.9 & 27.7 & \cellcolor{lred}20.8 & 22.5 & 
  27.5 & 43.4 & \cellcolor{lred}28.7 & 33.2   
  \\
  \rowcolor{lgreen}
  \bf{$\boldsymbol{\mu}$-MoE} &
  {16.4} & \bf{22.3} &  \bf{17.6} & \bf{18.8} & 
  \bf{18.0} & \bf{24.9} &  \bf{19.1} & \bf{20.7} & 
  \bf{23.1} & \bf{33.2} &  \bf{23.9} & \bf{26.7}  
  \\
  \toprule
  \multicolumn{13}{c}{OPT-2.7B (WT2: 12.5, PTB: 18.0, C4: 14.3, Avg: 14.9)}
  \\
  \midrule
  Magnitude Prune & 
  21.8 & 33.9 & 19.8 &  25.2 & %
  119.8 & 172.0 & 57.3 & 116.4 &  %
  4282.4 & 4667.9 & 2263.1 & 3737.8 %
  \\
  Wanda (WT2 Calib)  & 
  \cellcolor{lred} 13.0 & 19.6 & 16.0 & 16.2 & 
  \cellcolor{lred} 14.0 & 22.5 & 18.6 & 18.4 & 
  \cellcolor{lred} \bf{18.4} & 34.0 & 27.2 & 26.6  
  \\
  Wanda (PTB Calib)  & 
  13.2 & \cellcolor{lred}18.8 & 16.6 & 16.2 & 
  15.4 & \cellcolor{lred}20.4 & 19.7 & 18.5 & 
  26.1 & \cellcolor{lred}\bf{27.2} & 34.3 & 29.2 
  \\
  Wanda (C4 Calib) &
  \bf{12.9} & 19.1 &\cellcolor{lred} 15.1 & 15.7 & 
  14.5 & 21.9 & \cellcolor{lred}16.6 & 17.7 & 
  20.3 & 33.9 & \cellcolor{lred}22.2 & 25.5   
  \\
  \rowcolor{lgreen}
  \bf{$\boldsymbol{\mu}$-MoE} &
  {13.1} & \bf{18.6} & \bf{14.8} & \bf{15.5} & 
  \bf{13.8} & \bf{19.9} & \bf{15.6} & \bf{16.4} & 
  {18.5} & {31.6} & \bf{20.9} & \bf{23.6}   
  \\
  \toprule
  \multicolumn{13}{c}{OPT-6.7B (WT2: 10.9, PTB: 15.8, C4: 12.7, Avg: 13.1)}
  \\
  \midrule
  Magnitude Prune & 
  16.3 & 23.9 & 17.0 &  19.1 & %
  532.2 & 281.6 & 257.4 & 357.1 &  %
  9490.4 & 6743.4 & 6169.1 & 7467.6 %
  \\
  Wanda (WT2 Calib)  & 
  \cellcolor{lred} 11.0 & 17.2 & 14.2 & 14.2 & 
  \cellcolor{lred} 12.0 & 19.0 & 16.3 & 15.8 & 
  \cellcolor{lred} 15.1 & 25.0 & 22.8 & 21.0  
  \\
  Wanda (PTB Calib)  & 
  11.2 & \cellcolor{lred}16.3 & 14.6 & 14.0 & 
  13.6 & \cellcolor{lred}17.1 & 17.6 & 16.1 & 
  19.4 & \cellcolor{lred}20.6 & 25.8 & 21.9 
  \\
  Wanda (C4 Calib) &
  \bf{10.9} & 16.4 &\cellcolor{lred} 13.3 & 13.5 & 
  11.9 & 17.9 & \cellcolor{lred}14.3 & 14.7 & 
  15.3 & 23.6 & \cellcolor{lred}18.2 & 19.0   
  \\
  \rowcolor{lgreen}
  \bf{$\boldsymbol{\mu}$-MoE}  &
  {11.1} & \bf{16.1} &  \bf{13.0} & \bf{13.4} & 
  \bf{11.7} & \bf{16.7} &  \bf{13.5} & \bf{14.0} & 
  \bf{13.7} & \bf{19.7} &  \bf{15.7} & \bf{16.4}  
  \\
  \toprule
  \multicolumn{13}{c}{OPT-13B (WT2: 10.1, PTB: 14.5, C4: 12.1, Avg: 12.2)}
  \\
  \midrule
  Magnitude Prune & 
  59.8 & 78.4 & 44.5 &  60.9 & %
  2960.9 & 5406.3 & 3432.5 & 3933.2 &  %
  112900.6 & 28381.4 & 13734.1 & 51672.0 %
  \\
  Wanda (WT2 Calib)  & 
  \cellcolor{lred} 10.7 & 15.8 & 13.6 & 13.3 & 
  \cellcolor{lred} 12.0 & 18.7 & 15.7 & 15.5 & 
  \cellcolor{lred} 15.5 & 25.3 & 20.7 & 20.5 
  \\
  Wanda (PTB Calib)  & 
  10.9 & \cellcolor{lred}15.2 & 14.2 & 13.4 & 
  13.4 & \cellcolor{lred}16.8 & 17.4 & 15.9 & 
  20.6 & \cellcolor{lred}20.5 & 24.6 & 21.9 
  \\
  Wanda (C4 Calib) &
  10.9 & 15.2 &\cellcolor{lred}14.2 & 13.4 & 
  13.4 & 16.8 & \cellcolor{lred}17.4 & 15.9 & 
  20.6 & 20.5 & \cellcolor{lred}24.6 & 21.9   
  \\
  \rowcolor{lgreen}
  \bf{$\boldsymbol{\mu}$-MoE} &
  \bf{10.6} & \bf{15.0} &  \bf{12.3} & \bf{12.7} & 
  \bf{11.5} & \bf{16.4} &  \bf{12.9} & \bf{13.6} & 
  \bf{14.3} & \bf{20.2} &  \bf{14.6} & \bf{16.4}   
  \\
  \bottomrule
\end{tabular}
\end{table*}

\begin{table*}[t]
\centering
\caption{Accuracy in percent ($\uparrow$) on ScienceQA dataset of LLaVA-7B model with different compression methods for $40\%$--$60\%$ active weights.
Question
subjects: natural science (NAT); social science (SOC); language science (LAN).
Context modality: text (TXT);
image (IMG); or no context (NO).
Grades: 1--6 (G1-6); 7--12 (G7-12).
Wanda and SpargeGPT use TextVQA for calibration.
}
\label{tab:sqa_LLaVA}
\small
\newcolumntype{g}{>{\columncolor{lblue}}r}
\begin{tabular}{lc rrr rrr rr g}
  \toprule
   & & \multicolumn{3}{c}{Subject} & \multicolumn{3}{c}{Context Modality} & \multicolumn{2}{c}{Grades} \\
  \cmidrule(lr){3-5}
  \cmidrule(lr){6-8}
  \cmidrule(lr){9-10}
  Method & Active Weights &
  NAT & SOC & LAN &
  TXT & IMG & NO &
  G1-6 & G7-12 &
  Avg \\
  \midrule
  Original full & 100\% &
  72.47 & 69.18 & 65.73 & 
  73.51 & 68.82 & 65.99 &
  72.72 & 65.19 & 70.03
  \\
  \midrule 
  Magnitude Prune & 60\% &
  65.80 & 63.33 & 55.82 & 
  66.96 & 64.15 & 56.03 &
  66.34 & 56.16 & 62.70 \\
  SparseGPT & 60\% & %
  67.05 & \bf{65.47} & 55.91 &
  68.52 & \bf{66.73} & 57.21 & 
  66.12 & 59.72 & 63.83 \\
  Wanda & 60\% & %
  67.79 & 63.10 & 63.54 &
  67.94 & 63.31 & 62.79 & 
  67.66 & 60.71 & 65.17 \\
  \rowcolor{lgreen}
  \bf{$\boldsymbol{\mu}$-MoE} &  60\% &
  \bf{68.56} & {65.35} & \bf{65.73} &
  \bf{69.94} & {64.35} & \bf{65.09} &
  \bf{69.35} & \bf{63.22} & \bf{67.15} \\
  \midrule 
  Magnitude Prune & 50\% &
  40.63 & 46.01 & 39.91 & 
  37.05 & 36.94 & 43.90 &
  45.70 & 34.15 & 41.57 \\
  SparseGPT & 50\% & %
  55.02 & 48.14 & 53.55 &
  55.43 & 54.19 & 52.96 & 
  54.92 & 50.10 & 53.20 \\
  Wanda & 50\% & %
  59.33 & 56.81 & \bf{56.09} &
  60.07 & 56.32 & \bf{56.03} & 
  60.54 & 53.33 & 57.96 \\
  \rowcolor{lgreen}
  \bf{$\boldsymbol{\mu}$-MoE} &  50\% &
  \bf{63.23} & \bf{59.17} & {53.45} &
  \bf{64.37} & \bf{59.69} & {54.43} &
  \bf{63.36} & \bf{53.53} & \bf{59.84} \\
  \midrule 
  Magnitude Prune & 40\% &
  0.31 & 0.22 & 0.00 & 
  0.24 & 0.25 & 0.21 &
  0.18 & 0.26 & 0.21 \\
  SparseGPT & 40\% & %
  42.81 & 27.90 & 40.36 &
  43.60 & 37.08 & 38.40 & 
  40.16 & 37.05 & 39.05 \\
  Wanda & 40\% & %
  32.99 & 28.23 & 34.73 &
  30.16 & 31.68 & 35.47 & 
  33.22 & 31.05 & 32.45 \\
  \rowcolor{lgreen}
  \bf{$\boldsymbol{\mu}$-MoE} &  40\% &
  \bf{45.16} & \bf{35.21} & \bf{37.64} &
  \bf{44.62} & \bf{37.43} & \bf{40.07} &
  \bf{43.28} & \bf{37.24} & \bf{41.12} \\  \bottomrule
\end{tabular}
\end{table*}
\section{Expriments}

\paragraph{Experiments Setup} 

We conduct experiments for LLM benchmarks to evaluate the effectiveness of our method.
Our experiments are based on the same setting of SparseLLM~\cite{bai2024sparsellm} and their code base\footnote{\url{https://github.com/BaiTheBest/SparseLLM}}.
Following existing work~\cite{sun2023simple}, we compress all linear layers in LLM transformers to the target compression ratio.

For LLM experiments, we first consider the OPT model family~\cite{zhang2022opt} as it provides a wide range of model scales from 125M to 175B. 
We measure perplexity score for three popular benchmarks: raw-WikiText2 (WT2)~\cite{merity2016pointer}; the Penn Treebank (PTB)~\cite{marcus1994penn}; and C4~\cite{raffel2020exploring}.

We also analyze visual tasks for the LLaVA-7B model~\cite{liu2023LLaVA}, 
which consists of a language transformer based on Vicuna and a vision transformer tower.
We use the official code base\footnote{\url{https://github.com/haotian-liu/LLaVA}} to
evaluate the capability of the multi-modal answer reasoning for two benchmarks: ScienceQA~\cite{lu2022learn}; and TextVQA~\cite{singh2019towards}.
ScienceQA contains 21K vision-language multi-choice questions for three subjects: natural, social, and language science.
Some fractions of questions have image and/or text contexts, and the problem levels range from grades 1 to 12.
TextVQA makes LLMs to read and reason about text in images to answer visual reasoning questions for 28K images.

\paragraph{Impact of Model Size}

We first look into the impact of LLM model sizes in \cref{tab:perp_opt}, where perplexity of OPT models at active weight ratios of $60$--$40$\% are listed over 125M through 13B scales.
The perplexity results of the original full-parameter LLM models are reported next to the names of the models in the table.
Magnitude-based pruning is poor compared to offline Wanda pruning.
While Wanda is relatively robust over different calibration and test dataset, mismatched calibration often suffers a marginal loss.
Online Wanda pruning at $\mu$-MoE is found to be best for most cases across LLM sizes and compression ratios.

\begin{table}[t]
\centering
\caption{Accuracy in percent ($\uparrow$) on TextVQA dataset of LLaVA-7B model with different compression methods at 40--60\% active weights. 
Full-weight accuracy is 61.32\%.
Wanda and SparseGPT use ScienceQA for calibration.}
\label{tab:vqa}
\begin{tabular}{lrrr}
\toprule
Active Weights & 60\% & 50\% & 40\% \\
\midrule
Magnitude Prune & 
54.12  & %
45.56  & %
24.62 \\
SparseGPT & 
53.37  & %
47.42  & %
28.27 \\
Wanda & 
55.80  & %
52.36  & %
39.27 \\
\rowcolor{lgreen}
\bf{$\boldsymbol{\mu}$-MoE} & 
\bf{57.16}  & %
\bf{54.65}  & %
\bf{46.97} \\
\bottomrule
\end{tabular}
\end{table}

\paragraph{Multi-Modal Reasoning Capability}
We next show the accuracy of the LLaVA-7B model for the ScienceQA multi-modal reasoning benchmark in \cref{tab:sqa_LLaVA}.
SparseGPT and Wanda use TextVQA as the offline calibration data.
It is verified that our $\mu$-MoE can outperform offline pruning methods across diverse reasoning problems over most subjects, contexts, and grades.
Similar trends can be seen in another visual reasoning benchmark in \cref{tab:vqa}, where accuracy results for TextVQA are listed.
Here, Wanda and SparseGPT use ScienceQA as the calibration dataset.
In all experiments, $\mu$-MoE achieves better average performance over state-of-the-art baselines, especially for cases with fewer active weights.
The results suggest that online dynamic pruning can realize a task-agnostic MoE by adapting to every prompt given at test time.

\begin{table}[t]
\centering
\caption{Complexity of OPT-17B models with $\mu$-MoE.}
\label{tab:flops}
\setlength{\tabcolsep}{4pt}
\begin{tabular}{r r r r r}
\toprule
Active Weights & \multicolumn{2}{c}{FLOPs} & \multicolumn{2}{c}{MACs} \\
\midrule
100\% & 
3.29T & \progressbar{100}[1.2][red!70] & 
1.64T  & \progressbar{100}[1.2][red!70] \\
80\% & 3.21T 
& \progressbar{97.57}[1.2][red!70] & 
1.33T & \progressbar{81.10}[1.2][red!70]\\
60\% & 
2.55T & \progressbar{77.51}[1.2][red!70] & 
999G & \progressbar{60.91}[1.2][red!70] \\
40\% & 
1.90T & \progressbar{57.75}[1.2][red!70] & 
671G & \progressbar{40.91}[1.2][red!70] \\
20\% & 
1.24T & \progressbar{37.69}[1.2][red!70] & 
342G & \progressbar{20.85}[1.2][red!70] \\
\bottomrule
\end{tabular}
\end{table}

\paragraph{Computational Complexity}

We finally show the complexity analysis in \cref{tab:flops} for the OPT-17B models using our $\mu$-MoE dynamic pruning, based on the \texttt{calflops}\footnote{\url{https://pypi.org/project/calflops/}} library.
We included counts of floating point operations (FLOPs) and multiply-accumulate operations (MACs) for $\ell_2$-norm, top-$\rho$ value search, and comparators for instant Wanda pruning.
We use the token length of 128 for the analysis.
We found that the runtime complexity, especially in MACs, is almost proportional to the number of active weights.

\section{Conclusion}
We proposed a test-time pruning to realize mixture of micro-experts: $\mu$-MoE.
With the low complexity of Wanda pruning, online dynamic activation of massive single-parameter micro-experts became feasible for every prompt.
We demonstrated that $\mu$-MoE outperforms offline static pruning over several LLM benchmarks.
Studying the effect of fine-tuning for $\mu$-MoE is an interesting direction for future work.

\bibliography{ref}
\bibliographystyle{icml2025}

\newpage
\appendix
\onecolumn

\section{Related Work}

\paragraph{Model Compression}
The field of model compression for LLMs has aimed at mitigating the substantial computation and memory requirements~\cite{zhu2024survey, yuan2024llm}. 
Such methods primarily fall into four categories: weight quantization~\cite{lin2024awq, frantar2022gptq, wang2024q}, network pruning~\cite{lecun1989optimal, hassibi1993optimal, frantar2023sparsegpt, bai2024sparsellm},
knowledge distillation~\cite{hsieh2023distilling, deepseekai2025deepseekr1incentivizingreasoningcapability, hwang2024pc}, and rank reduction~\cite{yuan2023asvd, liu2024deepseek, hwang2024pc, saxena2024eigen, saha2024compressing}.

\paragraph{Static Pruning}
Model pruning~\cite{blalock2020state} methods generate a fixed reduced-parameter network. 
For example, weight pruning includes magnitude pruning~\cite{han2015deep}, pruning-aware retraining~\cite{frankle2018lottery} and activation-aware pruning~\cite{williams2023impact}.
SparseGPT~\cite{frantar2023sparsegpt} uses layer-wise optimal brain surgeon~\cite{dong2017learning, hassibi1993optimal, lecun1989optimal}, and SparseLLM~\cite{bai2024sparsellm} extends with joint multilayer perceptron (MLP) compression.
Wanda~\cite{sun2023simple} (as further discussed in \cref{app:wanda}) greatly simplifies the pruning mechanism, and has been extensively adopted for LLM post-training compression.
LLM pruner~\cite{ma2023llm} studied task-agnostic structured pruning.
\cite{bansal2022rethinking, liu2023deja, voita2023neurons} have demonstrated the existence of prompt-dependent and task-specific sparsity in LLMs.

\paragraph{Dynamic Pruning}

Dynamic networks~\cite{lin2017runtime, liu2018dynamic, hua2019channel, gao2018dynamic, chen2019self} selectively execute a subset of modules at inference time based on input samples. 
Typically, module selections are based on reinforcement learning or gating networks.
Adaptive dropout~\cite{ba2013adaptive, yang2025dynamic} is regarded as a fine-grained dynamic network.

\paragraph{Mixture of Experts (MoE)}

MoE~\cite{jiang2024mixtral, lin2024moe, liu2024deepseek} dynamically selects a subset of experts from a large pool.
Instead of using LLM experts, fine-grained MoE~\cite{krajewski2024scaling, xieautomated} uses relatively small experts..
The success and widespread use of parameter-efficient fine-tuning (PEFT)~\cite{hu2022lora, chen2024superlora, edalati2022krona, yeh2023navigating, bershatsky2024lotr, liu2023loda, koike2025quantum}, has enabled mixture of adapters~\cite{wu2024mixture, buehler2024x, wang2022adamix, zhang2024milora} to become a viable solution for task-agnostic MoE.

\section{Wanda: Efficient Activation-Aware Pruning}
\label{app:wanda}

\paragraph{Sorting Operation}

Wanda~\cite{sun2023simple} is a light-weight yet effective pruning method, suitable for online dynamic pruning.
The original algorithm uses a sorting operation to find the top-$k$ weights with high scores $S'_{i,j}$ per row:
\begin{lstlisting}
# W: (d', d), X: (d, Tc), kc=int((1-rho) * d)
S = W.abs() * X.norm(p=2, dim=-1)    # Score metric
_, idx = torch.sort(S, dim=-1)       # Sorting scores
pruned = idx[..., :kc]               # Select index having the kc smallest scores
W = torch.scatter(W, index=pruned, value=0)  # Zero-out weights
\end{lstlisting}
Here, $k_\mathrm{c} := (1-\rho)d$ is the complement of $k := \rho d$,
which means that $k$ corresponds to the number of active weights (micro-experts) per output neuron, whereas $k_\mathrm{c}$ corresponds to the number of inactive weights per output neuron.

\paragraph{Top-$k$ Search Operation}

Note that selecting the top-$k$ experts does not require a full sort operation.
Because the sorting operation is known to have a log-linear complexity order $\mathcal{O}[d' d\log(d)]$, an immediate alternative is to use the top-$k$ search operation instead of sorting as below:
\begin{lstlisting}
# W: (d', d), X: (d, Tc), kc=int((1-rho) * d)
S = W.abs() * X.norm(p=2, dim=-1)  
_, idx = torch.topk(S, dim=-1, k=kc, largest=False, sorted=False) # Select index having the kc smallest scores without sorting
W = torch.scatter(W, index=idx, value=0)
\end{lstlisting}
This should have a reduced theoretical complexity of $\mathcal{O}[d'd\log(d_\mathrm{c})]$, using the heap-based method.

\paragraph{The $k$th Value Search Operation}

Note that top-$k$ search can be also accomplished based on QuickSelect method searching for the $k$th largest value.
Hence, another option is to find the $k$th largest value to threshold scores:
\begin{lstlisting}
# W: (d', d), X: (d, Tc), kc=int((1-rho) * d)
S = W.abs() * X.norm(p=2, dim=-1)
val, _ = torch.kthvalue(S, dim=-1, k=kc)   # Find the kc-th smallest score
W = torch.where(S > val[:,None], W, 0)     # Activate weights whose scores are above it
\end{lstlisting}
Note that torch.kthvalue returns the $k_\mathrm{c}$th smallest value, not the largest value.
This should have the lowest theoretical complexity of $\mathcal{O}[d'd]$ on average.

\begin{figure}[h]
    \centering
    \begin{subfigure}[b]{0.32\linewidth}
        \includegraphics[width=\linewidth, trim=0 0 30 40, clip]{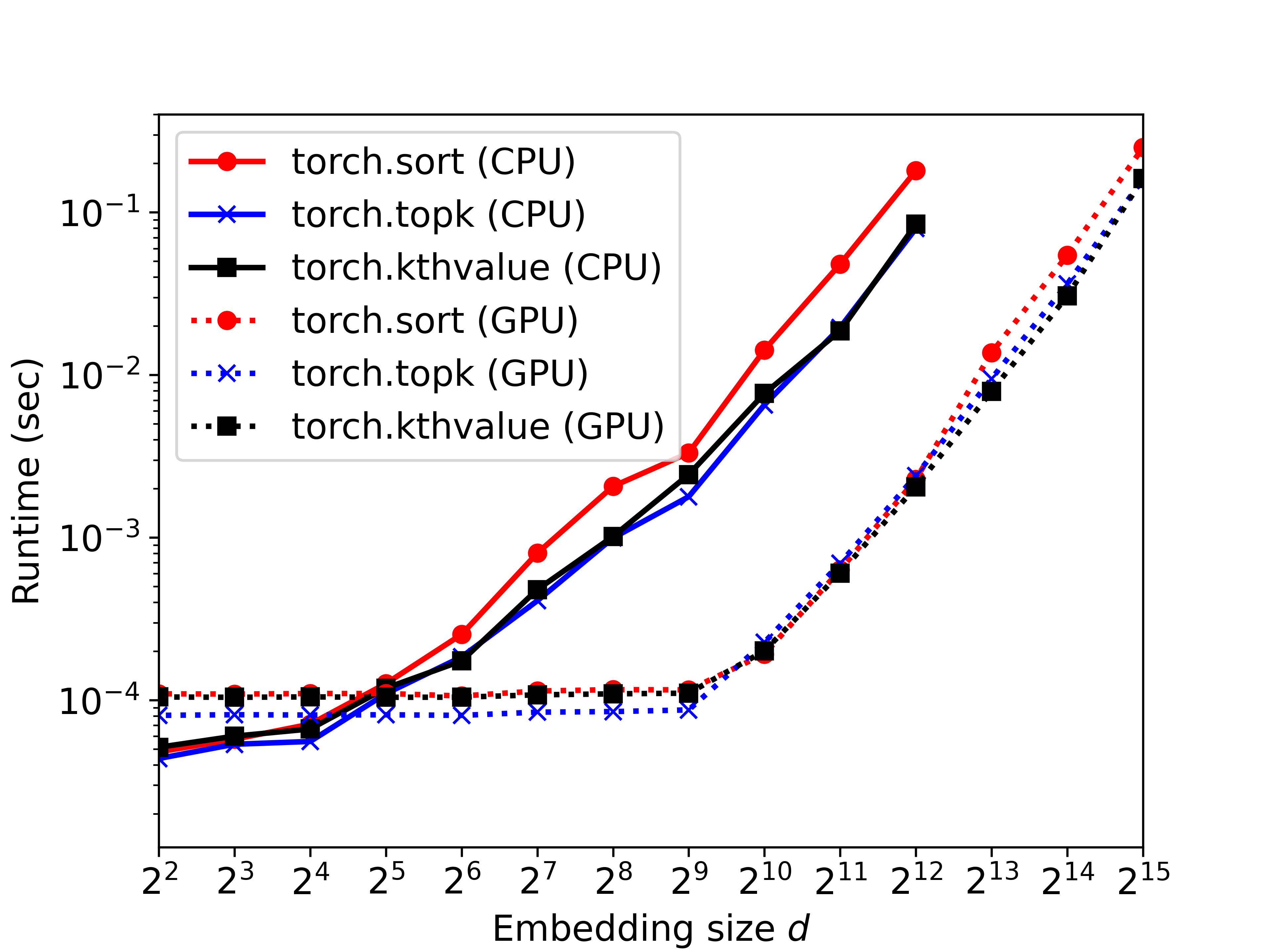}
        \caption{25\% Active Weights}
    \end{subfigure}
    \begin{subfigure}[b]{0.32\linewidth}
        \includegraphics[width=\linewidth, trim=0 0 30 40, clip]{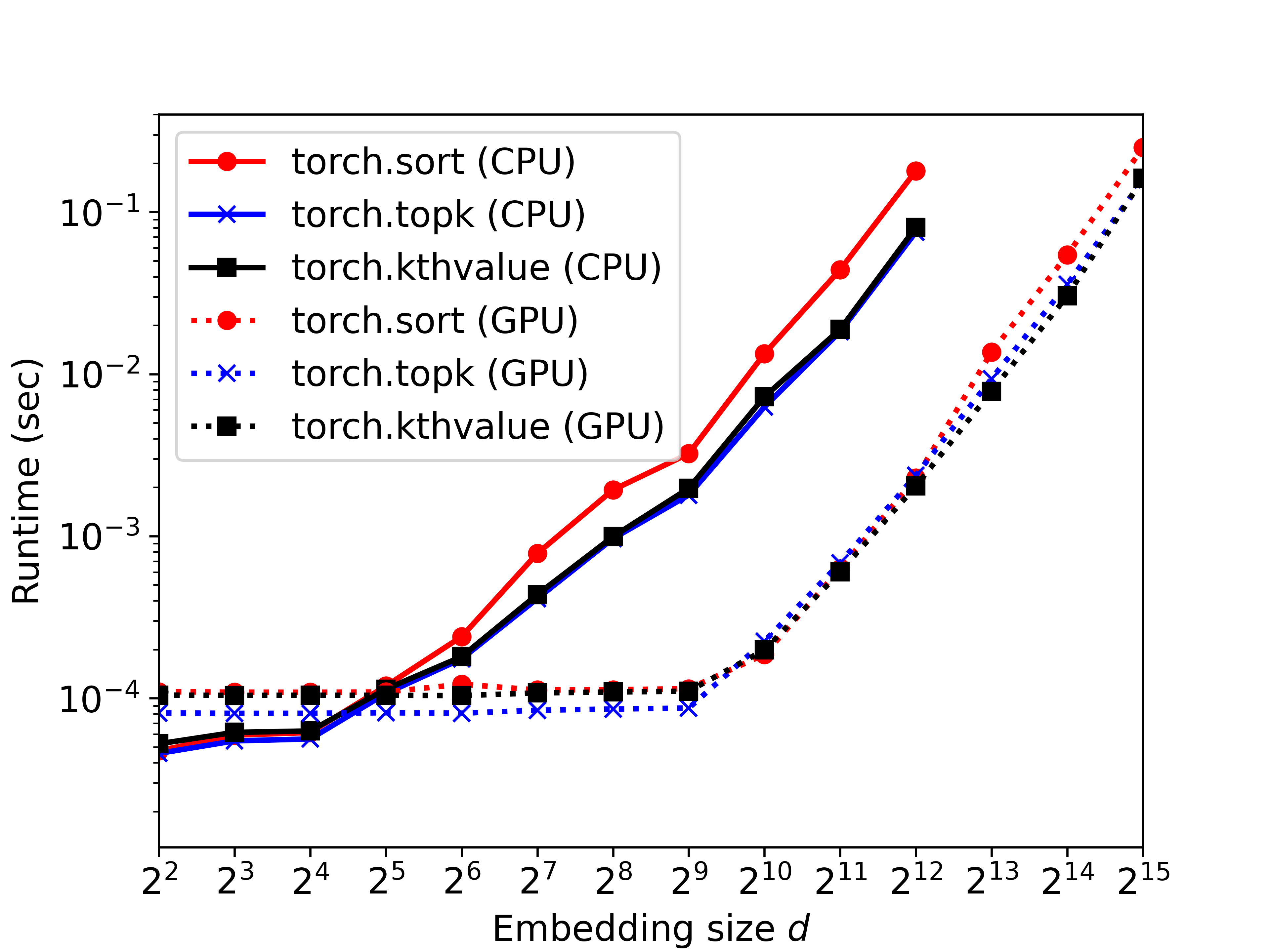}
        \caption{50\% Active Weights}
    \end{subfigure}
    \begin{subfigure}[b]{0.32\linewidth}
        \includegraphics[width=\linewidth, trim=0 0 30 40, clip]{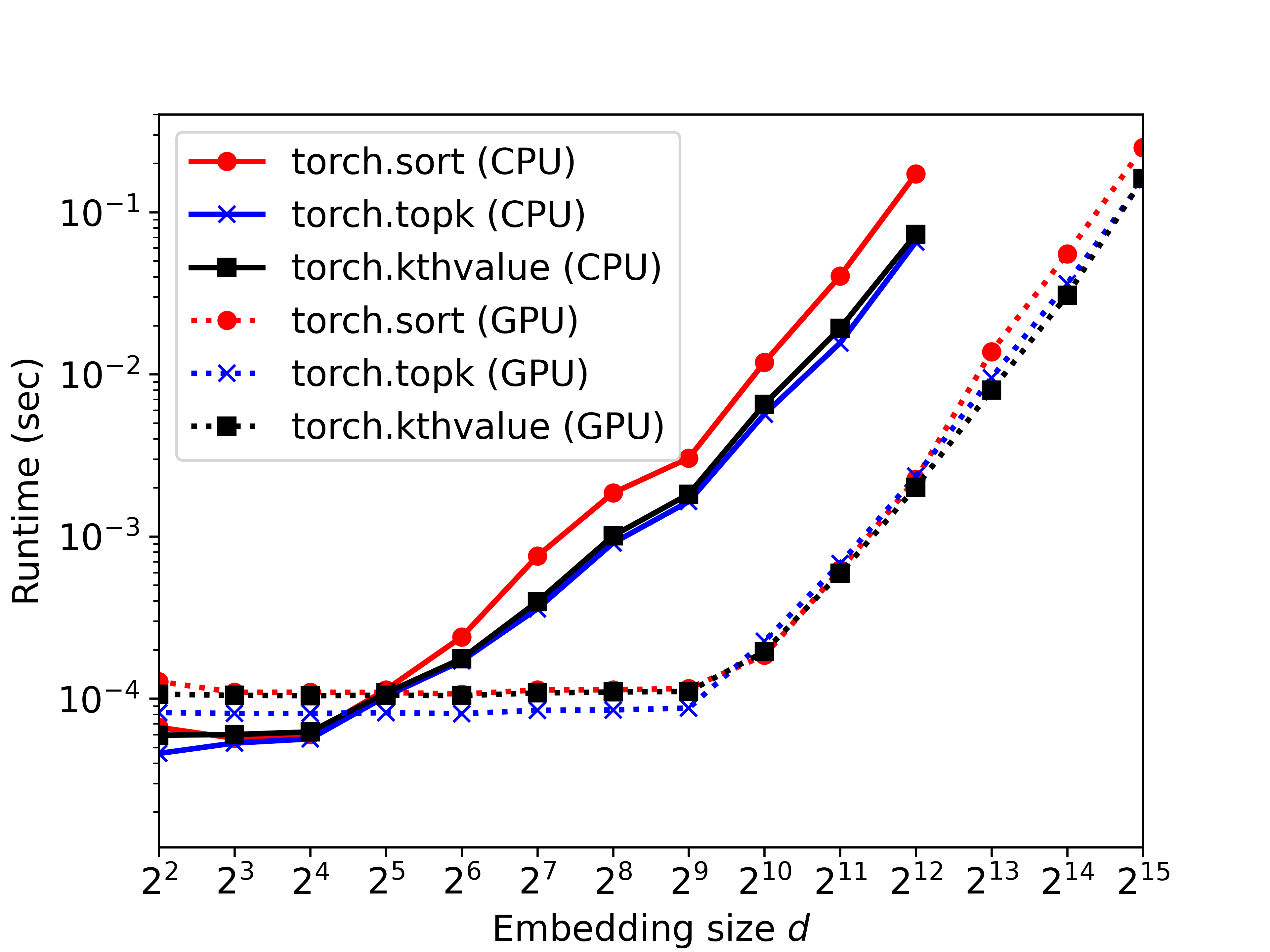}
        \caption{75\% Active Weights}
    \end{subfigure}
    \caption{Wanda pruning complexity based on torch.sort/topk/kthvalue on CPU and GPU at $\rho=0.25, 0.50, 0.75$.}
    \label{fig:topk}
\end{figure}

\paragraph{Runtime Analysis}
The practical complexity highly depends on its implementation on hardware.
An empirical experiment is shown in \cref{fig:topk}, where the average runtime for Wanda pruning over different embedding size $d$ is measured on Apple M1 CPU and NVIDIA A100 GPU.
It does not include the computation of linear affine transforms after weight pruning.
We see that torch.topk and torch.kthvalue can be moderately faster than torch.sort on the CPU, while there is no significant difference on the GPU.
Nevertheless, torch.topk and torch.kthvalue are found to be slightly advantageous for large weights on the GPU.
We also observe that the top-$k$ search computation is insensitive to the active weight ratio $\rho$.

\section{LLM Models}
\label{sec:model}

The Open Pre-trained Transformers (OPT)~\cite{zhang2022opt} is a suite of decoder-only pre-trained transformers ranging from 125M to 175B parameters. 
It was claimed that OPT-175B is comparable to GPT-3, while requiring only 1/7th the carbon footprint to develop.
\cref{tab:opt} shows model parameters for the OPT open LLM family.

\begin{table}[h]
\centering
\caption{OPT model parameters~\cite{zhang2022opt}}
\label{tab:opt}
\small
\begin{tabular}{rrrrrrl}
  \toprule
  Models & \# layers $L$& \# heads $h$ & hidden size $d$ & head dim $d_\mathrm{h}$ & $d_\mathrm{i}=4d$ & Huggingface ID \\
  \midrule
  125M   & 12 & 12 & 768 & 64 & 3072 & \href{https://huggingface.co/facebook/opt-125m}{facebook/opt-125m} \\
  350M   & 24 & 16 & 1024 & 64 & 4096 & \href{https://huggingface.co/facebook/opt-350m}{facebook/opt-350m} \\
  1.3B   & 24 & 32 & 2048 & 64 &8192 & \href{https://huggingface.co/facebook/opt-1.3b}{facebook/opt-1.3b}\\
  2.7B   & 32 & 32 & 2560 & 80 &10240 & \href{https://huggingface.co/facebook/opt-2.7b}{facebook/opt-2.7b}\\
  6.7B   & 32 & 32 & 4096 & 128 &16384 & \href{https://huggingface.co/facebook/opt-6.7b}{facebook/opt-6.7b}\\
  13B   & 40 & 40 & 5120 & 128 &20480 & \href{https://huggingface.co/facebook/opt-13b}{facebook/opt-13b} \\
  30B   & 48 & 56 & 7168 & 128&28672 & \href{https://huggingface.co/facebook/opt-30b}{facebook/opt-30b} \\
  66B   & 64 & 72 & 9216 & 128&36864 & \href{https://huggingface.co/facebook/opt-66b}{facebook/opt-66b} \\
  175B   & 96 & 96 & 12288 & 128 &49152 & --- \\
  \bottomrule
\end{tabular}
\end{table}

\section{LLM Experiment Results}

\begin{figure}[h]
    \centering
    \begin{subfigure}[b]{0.32\linewidth}
    \includegraphics[width=\linewidth, trim=0 0 30 40,clip]{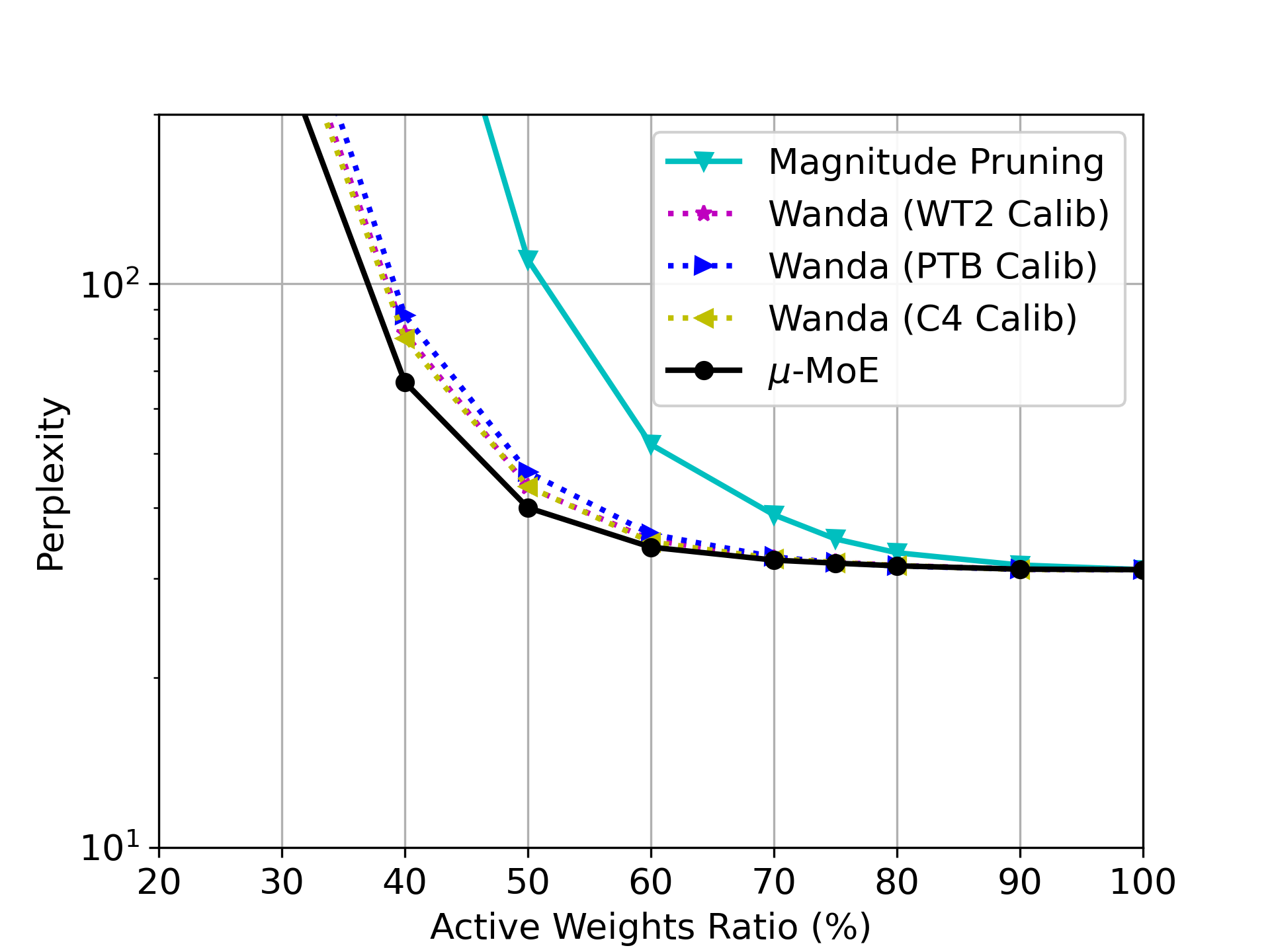}
    \caption{OPT-125M}
    \end{subfigure}
    \begin{subfigure}[b]{0.32\linewidth}
    \includegraphics[width=\linewidth, trim=0 0 30 40,clip]{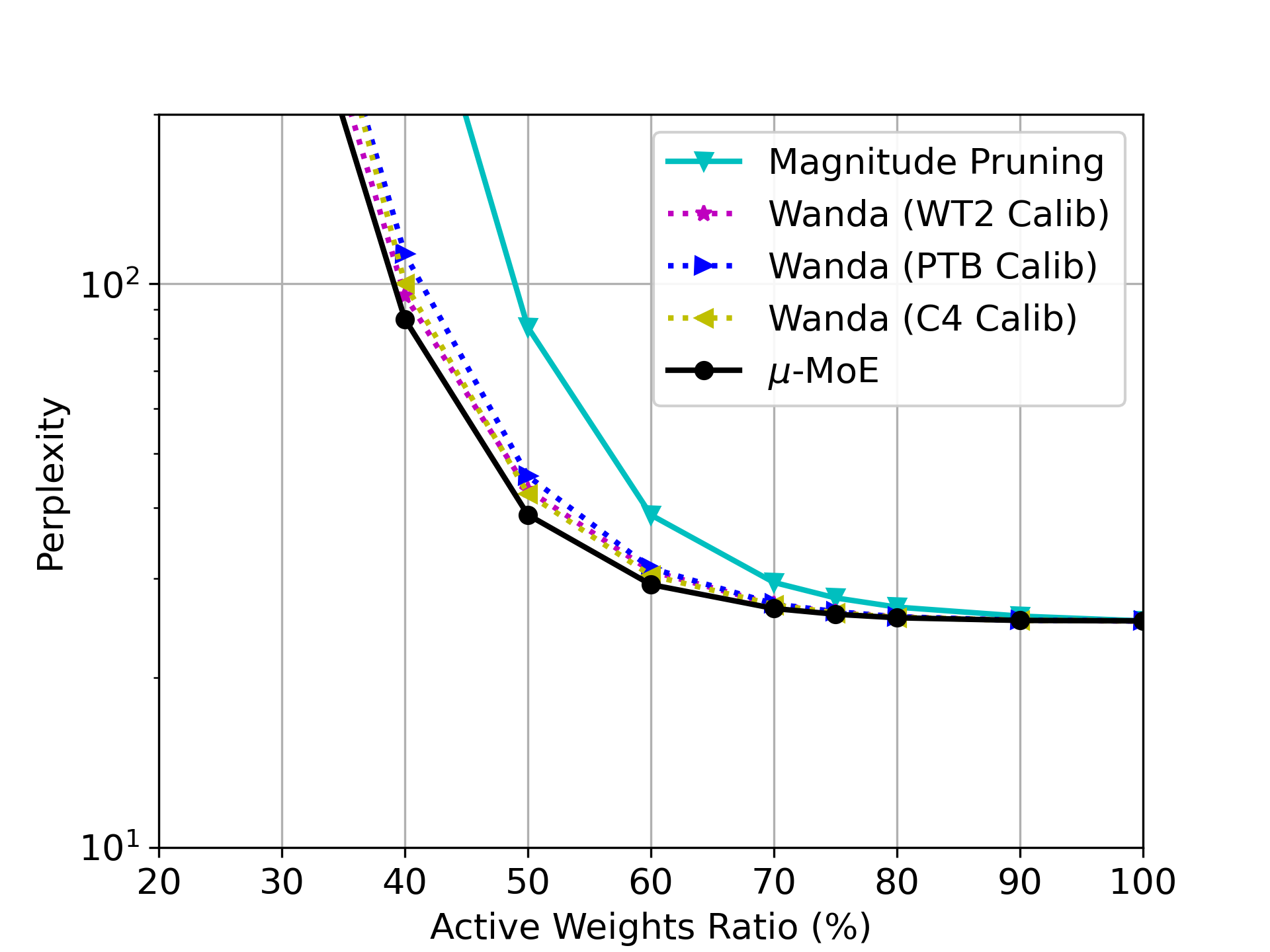}
    \caption{OPT-350M}
    \end{subfigure}
    \begin{subfigure}[b]{0.32\linewidth}
    \includegraphics[width=\linewidth, trim=0 0 30 40,clip]{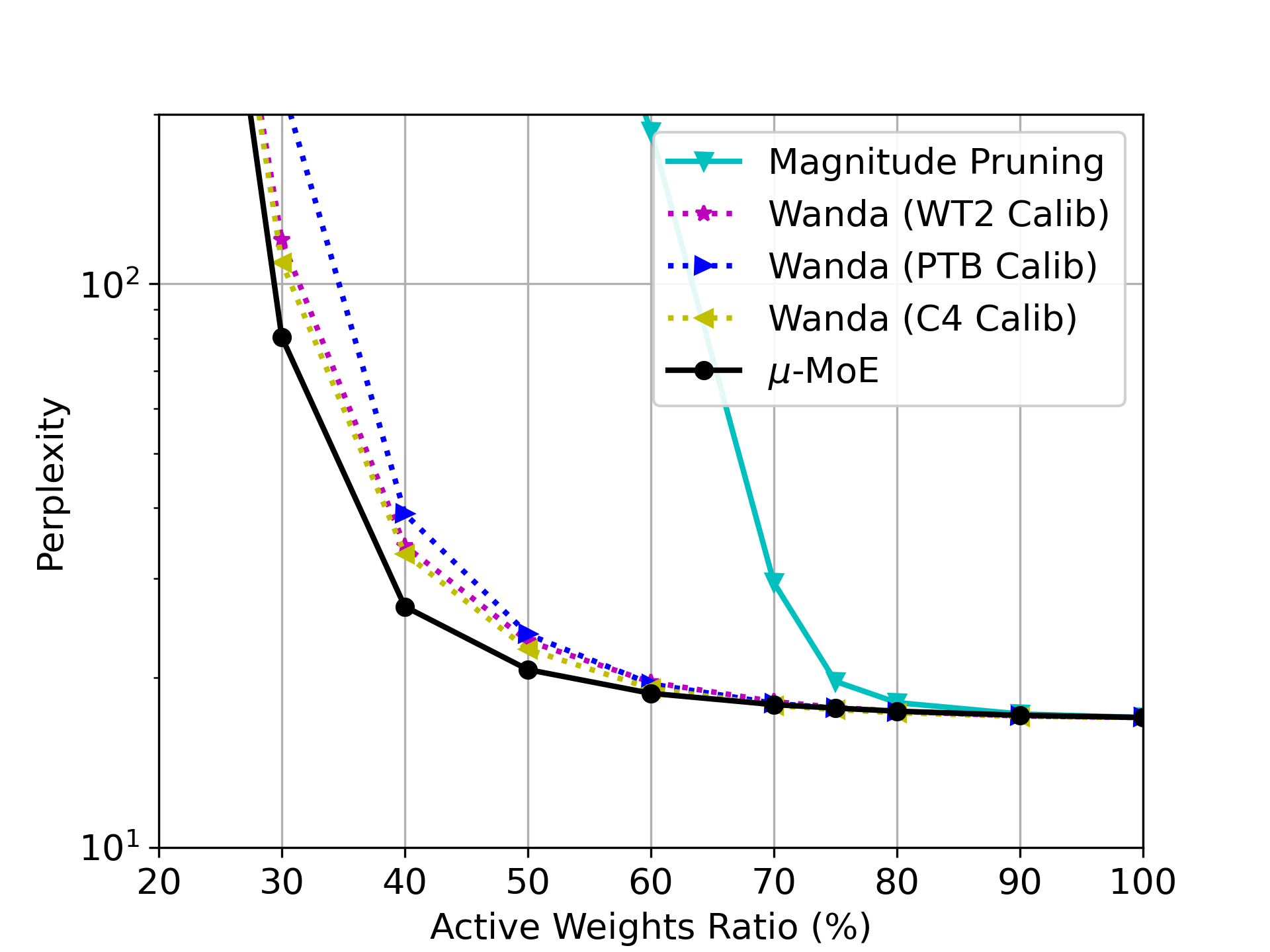}
    \caption{OPT-1.3B}
    \end{subfigure}
    \\
    \begin{subfigure}[b]{0.32\linewidth}
    \includegraphics[width=\linewidth, trim=0 0 30 40,clip]{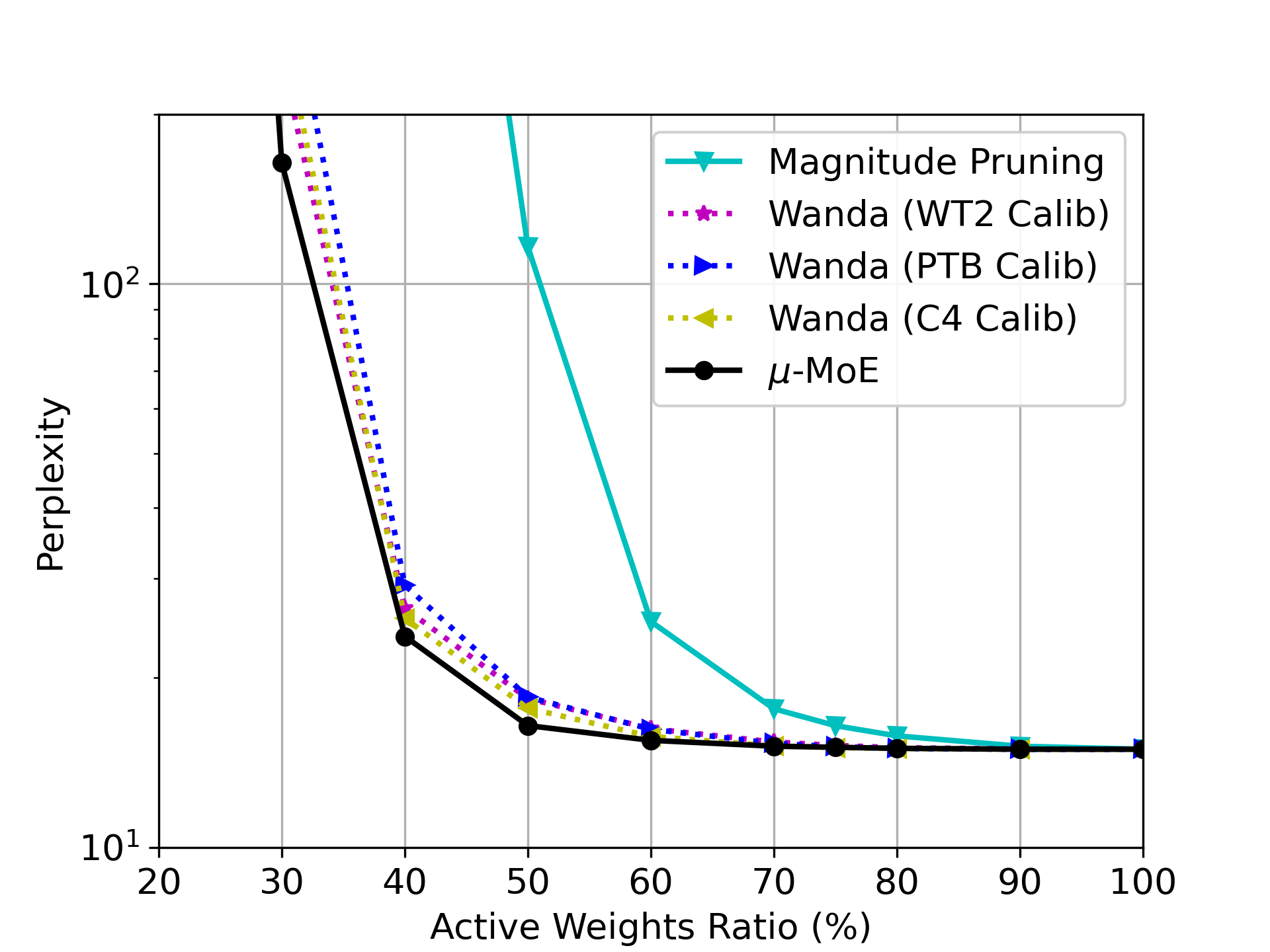}
    \caption{OPT-2.7B}
    \end{subfigure}
    \begin{subfigure}[b]{0.32\linewidth}
    \includegraphics[width=\linewidth, trim=0 0 30 40,clip]{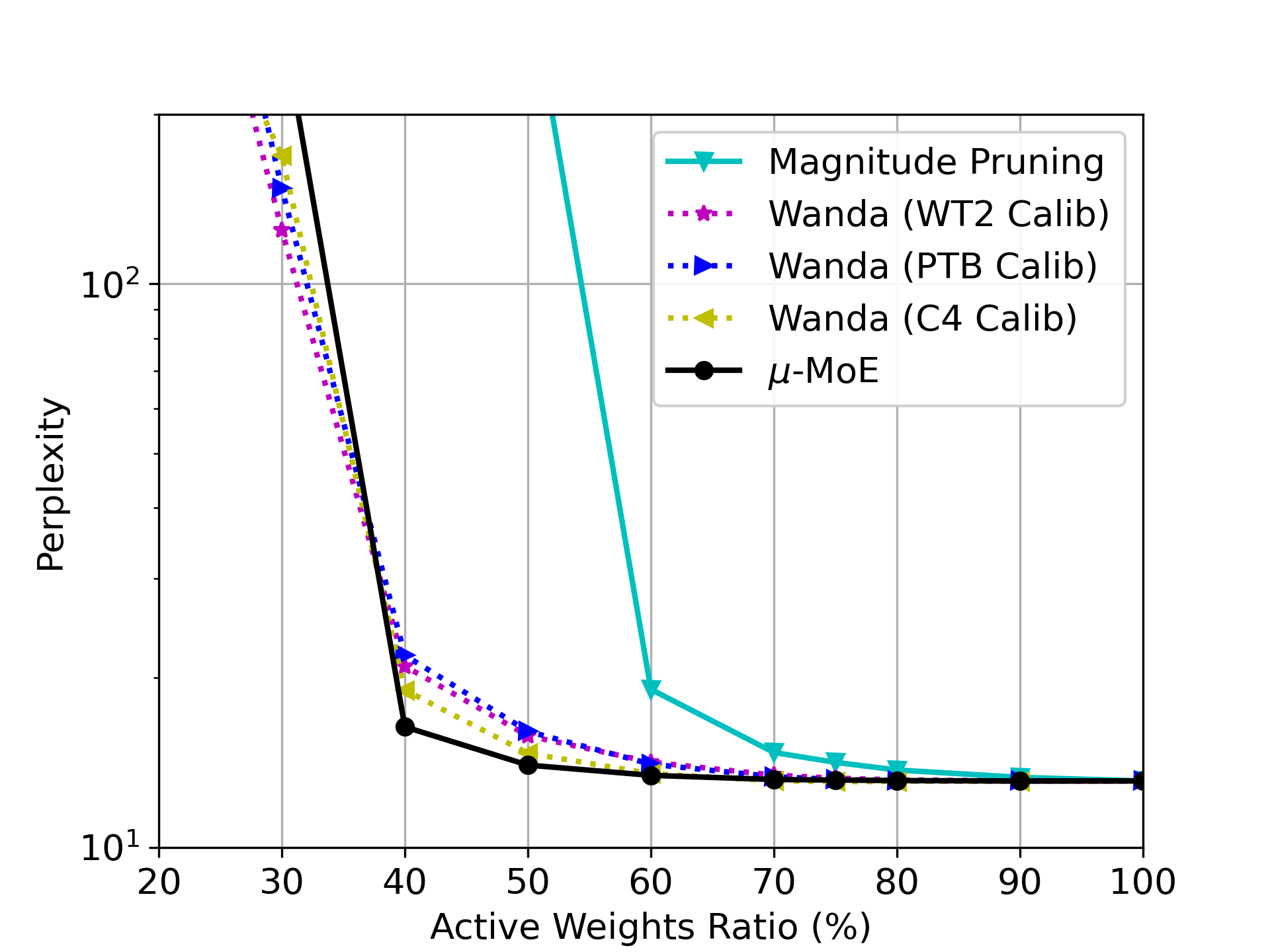}
    \caption{OPT-6.7B}
    \end{subfigure}
    \begin{subfigure}[b]{0.32\linewidth}
    \includegraphics[width=\linewidth, trim=0 0 30 40,clip]{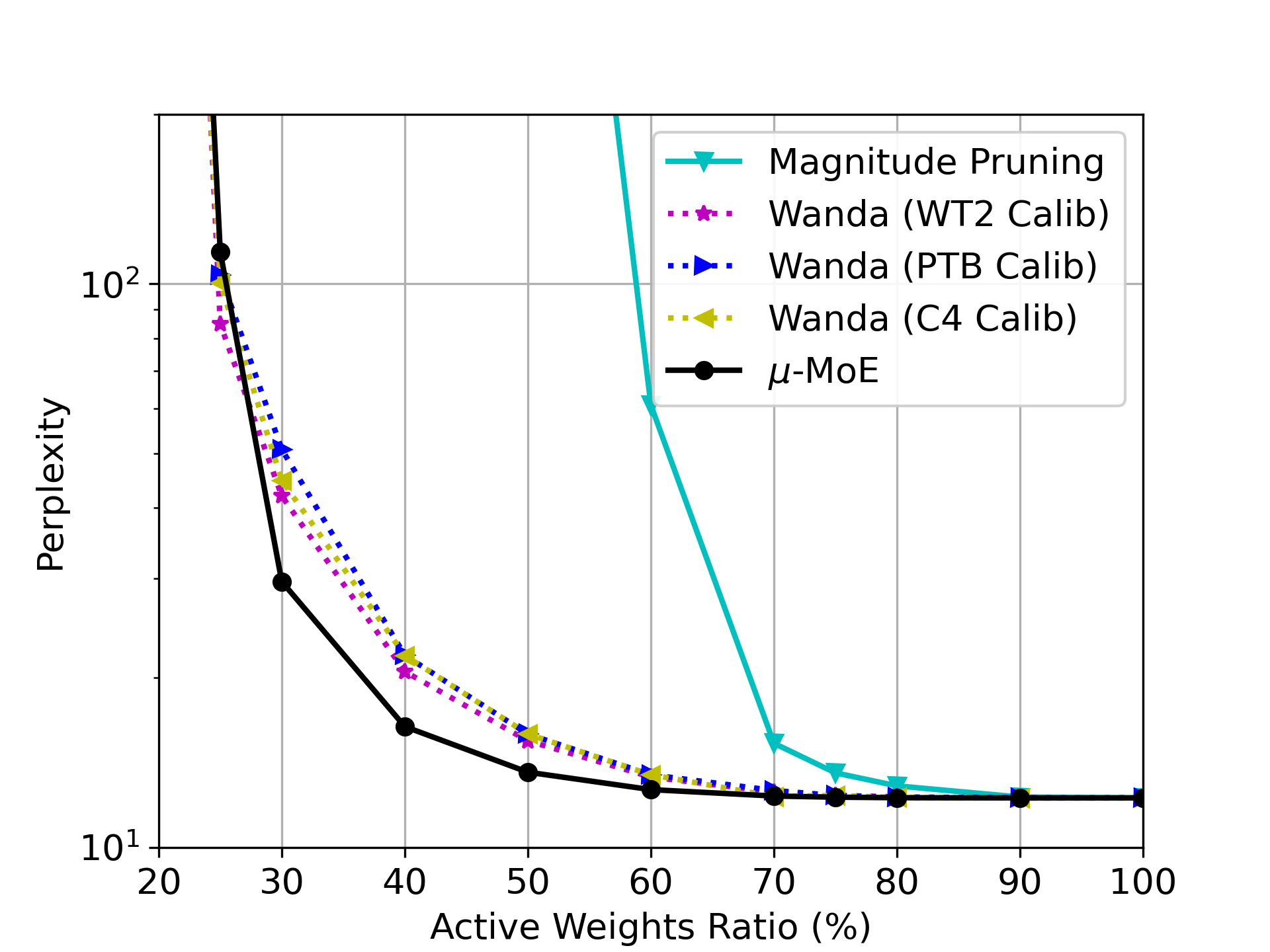}
    \caption{OPT-13B}
    \end{subfigure}
    \caption{Perplexity results averaged over WT2, PTB, and C4 datasets for compressed OPT models.}
    \label{fig:opt}
\end{figure}

\paragraph{Impact of Model Size}
\cref{fig:opt} plots the perplexity results averaged over the WT2, PTB, and C4 datasets for compressed OPT models of 125M, 1.3B, and 13B scales.
This partly corresponds to \cref{tab:perp_opt}, while including a wider range of compression ratios.
We can see that the magnitude pruning is poor and activation-aware pruning works well.
Online pruning with $\mu$-MoE can further improve the perplexity through prompt-wise adaptation, especially around 40\%.

\section{Datasets}
\paragraph{Wikitext-2 (WT2)}
The WikiText language modeling dataset~\cite{merity2016pointer} is a collection of over 100 million tokens extracted from the set of verified good and featured articles on Wikipedia. 
The dataset is available under the CC BY-SA-4.0 license.
The wikitext-2-raw-v1 contains 36{,}718, 3{,}760, and 4{,}358 samples for train, validation, and test splits, respectively.
We use \url{https://huggingface.co/datasets/mindchain/wikitext2}.

\paragraph{Penn Treebank (PTB)}
The English Penn Treebank (PTB) corpus~\cite{marcus1994penn}
is one of the most known and used corpus for the evaluation of models for sequence labeling. 
The dataset features a million words of 1989 Wall Street Journal material. 
We use \url{https://huggingface.co/datasets/ptb-text-only/ptb_text_only}.

\paragraph{C4}
C4~\cite{raffel2020exploring} is based on a colossal, cleaned version of Common Crawl's web crawl corpus.
This is release under the OCD-By license.
We consider a subset ``en'', containing 364{,}868{,}892 and 364{,}608 samples for train and validation splits, respectively, while we use the first shard for each split in \url{https://huggingface.co/datasets/allenai/c4}.

\paragraph{ScienceQA}
ScienceQA~\cite{lu2022learn} is collected from elementary and high school science curricula (i.e., grades 1 through 12), and contains 21{,}208 multimodal multiple-choice science questions. 
Out of the questions in ScienceQA, 10{,}332 (48.7\%) have an image context, 10{,}220 (48.2\%) have a text context, and 6{,}532 (30.8\%) have both. 
Most questions are annotated with grounded lectures (83.9\%) and detailed explanations (90.5\%). 
The lecture and explanation provide general external knowledge and specific reasons, respectively, for arriving at the correct answer. 
ScienceQA has rich domain diversity from three subjects: natural science, language science, and social science. 
ScienceQA features 26 topics, 127 categories, and 379 skills that cover a wide range of domains.
It contains 12{,}726, 4{,}241, and 4{,}241 samples for train, validation, and test splits in \url{https://huggingface.co/datasets/derek-thomas/ScienceQA}.
This is released under the CC BY-NC-SA 4.0 license.

\paragraph{TextVQA}

TextVQA~\cite{singh2019towards} requires VLM models to read and reason about text in images to answer questions about them. 
Specifically, models need to incorporate the new modality of text present in the images and reason over it to answer TextVQA questions. 
TextVQA dataset contains 45{,}336 questions over 28{,}408 images from the OpenImages dataset. 
We use \url{https://huggingface.co/datasets/facebook/textvqa}.
This is licensed under CC-BY-4.0.

\end{document}